# Distributed Intelligent System Architecture for UAV-Assisted Monitoring of Wind Energy Infrastructure

Serhii Svystun[1,*,†], Oleksandr Melnychenko[1,†], Pavlo Radiuk[1], Oleg Savenko[1] and Andrii Lysyi[1]

[1] *Khmelnytskyi National University, 11, Institutes str., Khmelnytskyi, 29016, Ukraine*

**Abstract**

With the rapid development of green energy, the efficiency and reliability of wind turbines are key to sustainable renewable energy production. For that reason, this paper presents a novel intelligent system architecture designed for the dynamic collection and real-time processing of visual data to detect defects in wind turbines. The system employs advanced algorithms within a distributed framework to enhance inspection accuracy and efficiency using unmanned aerial vehicles (UAVs) with integrated visual and thermal sensors. An experimental study conducted at the "Staryi Sambir-1" wind power plant in Ukraine demonstrates the system's effectiveness, showing a significant improvement in defect detection accuracy (up to 94%) and a reduction in inspection time per turbine (down to 1.5 hours) compared to traditional methods. The results show that the proposed intelligent system architecture provides a scalable and reliable solution for wind turbine maintenance, contributing to the durability and performance of renewable energy infrastructure.

## 1. Introduction

In recent decades, the growth of renewable energy sources, such as wind and solar power, has increased the importance of proper maintenance and efficient operation of these technologies. Wind turbines, a pivotal component in wind energy generation, pose unique challenges due to their substantial size and intricate mechanical structures [1]. At the same time, they tend to take different forms of degradation over time, including blade cracks and mechanical failures, which can compromise performance or lead to complete shutdowns. Failure to detect such issues promptly can result in significant repair costs and loss of energy production [2, 3], emphasizing the critical need for continuous and reliable monitoring systems [4].

Conventional inspection methods, such as manual visual assessments and ground-based equipment checks, often fall short in efficiency, speed, and cost-effectiveness. These approaches can be labor-intensive, time-consuming, and sometimes risky for personnel due to turbine components' elevated and exposed locations. The advent of unmanned aerial vehicles (UAVs) [5, 6], coupled with advancements in visual data processing technologies [7, 8], has opened new avenues for automated inspection solutions [9]. These technologies facilitate real-time identification of potential problems, thereby minimizing downtime and extending the operational lifespan of wind turbines [10].

However, existing automated systems often struggle with limitations like insufficient scalability, inadequate real-time data processing capabilities, and challenges in integrating multiple sensor types for a holistic assessment [11]. Hence, there is a need for an intelligent system capable of dynamically collecting and analyzing visual data on turbine defects, effectively integrating various sensors, and operating efficiently across expansive wind farms.

Addressing these challenges, this research presents a novel intelligent system architecture designed to enhance the dynamic collection and processing of visual data to identify defects in green energy infrastructures in real time, focusing on wind turbines. The proposed system leverages UAVs with integrated sensors, including visual and thermal cameras, and utilizes advanced algorithms to comprehensively evaluate turbine conditions. Its distributed architecture allows for the parallel operation of different components, such as UAVs, control devices, and sensors, ensuring seamless functionality in large-scale wind farm environments.

The primary objective of this study is to develop an advanced, automated monitoring solution that overcomes the limitations of current systems by improving efficiency, accuracy, and scalability in wind turbine inspections. By harnessing state-of-the-art UAV technology, sensor integration, and real-time data processing, the system aims to provide a more reliable and effective means of maintaining the health and performance of wind turbines. This contributes to the broader goal of sustaining and advancing renewable energy infrastructures for long-term viability.

## 2. Related works

Efforts to develop effective monitoring and inspection systems for wind turbines have intensified, focusing on early defect detection to mitigate unexpected failures and ensure continuous operation [12, 13]. Traditional methods, such as manual visual inspections [14] and ground-based equipment assessments [15], are increasingly considered inadequate due to inefficiency, high costs, and safety risks [16]. This has led to a growing interest in integrating advanced technologies like UAVs, sensor fusion, and deep learning (DL) [17] methods into inspection practices [18].

Several studies have explored the use of UAVs for wind turbine inspections. Sanati et al. introduced an automated UAV-based system utilizing visual and thermal imaging to detect defects like blade cracks [19]. While this method enhanced detection capabilities, it was limited by its reliance on specific imaging modalities, potentially missing defects detectable by other sensors. Shihavuddin et al. employed UAVs equipped with convolutional neural networks (CNNs) to improve surface defect detection [20]. Although this approach increased accuracy and explainability rates, it often depended on single-sensor data, which might not capture the full spectrum of potential defects.

Recently, different research groups have investigated multi-sensor fusion techniques to overcome the limitations of single-sensor reliance. Memari et al. integrated visual and thermal sensors to develop an intelligent system capable of detecting surface and internal defects in turbine blades [21]. This comprehensive approach improved reliability but introduced complexity in data integration and required sophisticated algorithms for effective processing. Similarly, Castelar Wembers et al. proposed a multi-sensor system combining LiDAR and infrared sensors to inspect turbine towers for structural defects [22]. However, this method demanded advanced equipment and increased computational resources.

Advancements in intelligent data processing have also been significant. Tian et al. presented a DL-based method using CNN and long short-term memory networks to analyze collected data and identify blade defects [23]. This technique improved detection accuracy but was computationally intensive, potentially hindering real-time application. Li et al. developed a DL model to predict turbine components' remaining useful life, enabling predictive maintenance and reducing sudden breakdown risks [24]. However, this model required extensive datasets and high computational power, which might only be feasible in some operational settings.

Distributed system architectures have been proposed to enhance scalability and efficiency. Jacobsen et al. suggested a distributed UAV network allowing multiple drones to inspect large wind farms simultaneously, significantly reducing data collection time [25]. Despite the efficiency gains, coordinating multiple UAVs presents challenges in communication and synchronization. Mokhtar et al. integrated cloud and edge computing into a distributed inspection system, facilitating real-time data processing and quicker decision-making [26]. Nevertheless, relying solely on cloud infrastructure may lead to sensitive data leaks and data latency issues.

In summary, while significant progress has been made in wind turbine inspection technologies, challenges persist in achieving comprehensive, real-time monitoring that is both scalable and efficient. Existing systems often face limitations such as dependency on single-sensor data, high computational demands of advanced algorithms, and complexities in coordinating distributed systems.

To address the abovementioned issues, this study aims to enhance the dynamic collection and real-time processing of visual data on wind turbine defects by developing an intelligent system that addresses these challenges. The scientific contributions of this study are as follows:

- The proposed intelligent system architecture combines data from various sensors, including the UAV's visual and thermal cameras, into a distributed structure, providing a comprehensive assessment of the monitoring process while maintaining scalability.
- By employing intelligent algorithms optimized for real-time processing, the system improves the accuracy and speed of defect identification without excessive computational requirements.
- The system utilizes a distributed network of UAVs with adaptive coordination strategies to ensure seamless operation across large wind farms, reducing inspection time and enhancing data collection efficiency.

These contributions aim to resolve existing issues in the field by providing a more effective and scalable solution for wind turbine monitoring and maintenance.

# 3. Proposed architecture of intelligent systems

The task of collecting visual data on defects in green energy objects within a three-dimensional space necessitates the development of a comprehensive system that integrates various tools with different functional purposes. Given the specific nature of the objects under investigation, it is advisable to utilize and manage geographically distributed resources. Communication support is provided through appropriate information and communication technologies. A key element of the system should be the provision of automated management of all components, which is essential for the rational use of the system by end-users. Consequently, the challenge arises to design the intelligent system $M_s$ to maximize the automation of data collection and analysis processes in real-time.

The components of the designed intelligent system $M_s$ are presented in the general structural diagram and include various elements. The diversity of system components and their distribution significantly impact the system's operation. Therefore, the primary tasks in designing it should focus on ensuring the system's integrity during operation and achieving maximum resource efficiency.

To design the intelligent system $M_s$, it is necessary to first identify the objects it targets. Primarily, these are green energy objects, particularly wind turbines and their components. These objects are characterized by complex structures and large sizes, which require a special approach to the collection and processing of visual data.

The objects of study are located within a distributed three-dimensional space $\mathbb{R}^3$. To ensure precise collection of visual data for further analysis of detected defects across the entire area, the study region is defined as a set of spatial zones $Z_i$, each including an individual wind turbine and its surroundings. For this purpose, we use the coordinates of the central points $C_i(x_i, y_i, z_i)$ of the turbines and the corresponding coverage radii $R_i$, which ensure full coverage of the entire structure.

Each spatial zone $Z_i$ can be represented as a sphere with center $C_i$ and radius $R_i$:

$$Z_i = \{P \in R^3 \mid |P - C_i| \leq R_i\}, \tag{1}$$

where $|P - C_i|$ is the Euclidean distance between point $P$ and the central point $C_i$.

The total coverage area $A$ can be expressed as the union of all spatial zones:

$$A = \bigcup_{i=1}^{n} Z_i. \tag{2}$$

The studied space $A$ is defined as a set of spatial vectors $V_i$, and the vector $V$, which describes the coordinates of the central point and the coverage radius for each wind turbine in the designed system $M_s$, is represented as follows:

$$V_i = \langle C_i(x_i, y_i, z_i), R_i \rangle, \quad i = 1, 2, \dots, N, \tag{3}$$

where $C_i(x_i, y_i, z_i)$ are the coordinates of the central point of the $i$-th wind turbine, $R_i$ is the coverage radius for this turbine, and N is the total number of turbines in the study area.

Thus, the area A can be represented as the set of all vectors $V_i$, which determine the coordinates and coverage radii: $A = \{V_i \mid i = 1, 2, \dots, N\}$. This set of vectors defines the complete study space in the designed system $M_s$. Therefore, we represent the spatial area in the form of a vector matrix:

$$M_v = \begin{pmatrix} v_{1,x1} & \cdots & v_{N_v,x1} \\ \vdots & \ddots & \vdots \\ v_{1,v_{3,3}} & \cdots & v_{N_v,v_{3,3}} \end{pmatrix}, \quad (4)$$

where $(v_{1,x1} \ldots v_{1,v_{3,3}})$ are the 12 coordinates of the first vector $(v_1)$ by formula (3).

Similarly, for the remaining $(N_v - 1)$ vectors. The values obtained from formula (4) are used as input data for the $M_s$ system. Its structural diagram is shown in Figure 1.

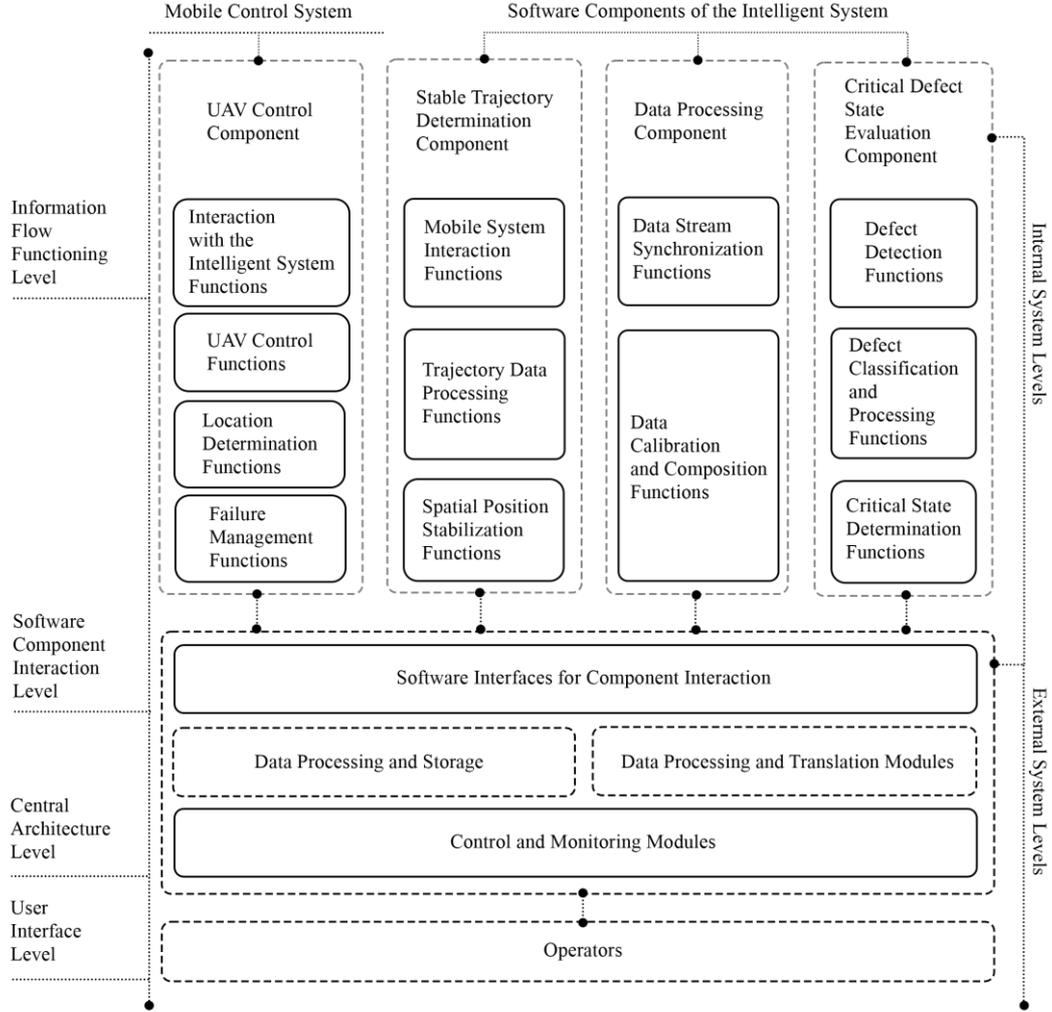

**Figure 1:** Structural diagram of the intelligent system $M_s$.

Since data collection requires parallel functioning, where different software components of the system operate simultaneously, and many components perform their tasks in parallel, it is advisable to design such an intelligent system $M_s$ as a distributed one. Implementing a distributed architecture allows multiple data processing and component management operations to be executed simultaneously, reducing delays and improving performance. A distributed system ensures scalability and reliability, which are critical for processing large volumes of data in real time and maintaining uninterrupted operation of all components.

The system's component structure includes UAVs, mobile control devices, computing complexes, maintenance tools, and multifunctional sensors operating in an automated mode.

The diagram shown in Figure 1 illustrates the architecture of intelligent systems for collecting visual data and assessing the condition of defects in green energy objects, particularly wind turbines. The system consists of several levels and includes both hardware and software components that are integrated to ensure automated data collection, processing, and analysis.

In terms of functionality, each software and hardware component are defined at any given moment by its state, described by its functional capabilities [27]. Thus, the state of an individual component can be represented as a vector: $S_{f,s}(t) = \{P_{f,s1}(t), P_{f,s2}(t), \ldots, P_{f,sn_{f,s}}(t)\}$, where $P_{f,s}(t)$ represents the parameters of the system components at time $t$, defining their functional capabilities, and $n_{f,s}$ is the number of such parameters. The set of all component states at a given time $t$ forms the overall state of the system:

$$S(t) = \bigcup_{f,s} S_{f,s}(t), \tag{5}$$

The system components interact through specific connections, which can be represented using graphs. In this graph, vertices represent the system components, while edges denote the connections between them. The inter-component interactions are captured by the graph $G_{f,s}$, where $f$ defines the functions representing the connections between components, and $s$ indicates the states of these components. Thus, the system's graph is defined as follows:

$$G_{f,s} = (V, E), \tag{6}$$

where $V$ denotes the set of vertices representing the system components, and $E$ represents the set of edges that define the connections between these components.

Each edge $e_{ij} \in E$ connects vertices $v_i$ and $v_j$, corresponding to the interaction between components $i$ and $j$. Each connection in the system can be described by a function that specifies how one component influences another. The set of all connections within the system is expressed as:

$$E = \{e_{ij}(S_i, S_j) = f_{ij}\left(P_{ik}(t), P_{jl}(t)\right) \mid \forall i, j, k, l\}, \tag{7}$$

where $f_{ij}$ represents the functions that describe how the parameters of one component influence those of another.

Thus, the intelligent system $M_s$ is characterized by its elements and the connections among them, defined as follows:

$$M_{s,Z} = \{S(t), E\}, \tag{8}$$

where $M_{s,Z}$ represents the system $M_s$, including $Z$, which refers to its architectural task according to Figure 1.

The intelligent system $M_s$ must provide the following functions:

- Organizing the preparation and connection of UAVs to the system.
- Creating a three-dimensional software space $A$.

- Determining the initial position of the hardware device relative to the object under study in space $A$.
- Managing the UAV's trajectory.
- Synchronizing and processing data in real time.
- Integrating with other management and monitoring systems.
- Defining the states of the system and the object under study.

The objective of the system $M_s$, as outlined in formula (8), is to represent all elements of the system, along with the processes and interactions occurring among them. Since the heterogeneous components of the intelligent system $M_s$ and the input data specified in formula (8) are not directly addressed, we define the system $M_s$ as follows to account for these components:

$$M_{S_Z} = \langle K_{S,Z,1}, K_{S,Z,2}, \ldots, K_{S,Z,nf,Z} \rangle, \tag{9}$$

where $nf_Z$ is the number of sets indicating various system components by their characteristic features, and $K_{S,Z,i}$ is a set whose elements are homogeneous components that comprise the system $M_s$, with $i = 1,2,\ldots,nf_Z$. The designation $Z$ in $M_{S_Z}$ for system description refers to its architectural definition, which includes the physical components.

The final output of the $M_s$ system is derived from the visual data gathered by external sensors and the intelligent processing applied to achieve the target task. According to formula (4), the operational three-dimensional space $A$ consists of a set of vectors $V_i$, so the data collected by the external sensors is represented in matrix form as follows:

$$M_{J_k} = \begin{bmatrix} d_{1,1} & \cdots & d_{Nd,1} \\ \vdots & \ddots & \vdots \\ d_{1,n} & \cdots & d_{Nd,n} \end{bmatrix}, \tag{10}$$

where $d_{i,j}$ denotes an element of the matrix $M_{J_k}$, representing a vector that contains information gathered by external sensors from the target areas within space $(i,j)$.

The information, collected by external sensors and stored in matrix $M_{J_k}$, is processed within the corresponding hardware module, and the processing results contribute to the final outcome. To formalize this, we define a function $F_{jk}$, which, for each element of the matrix $M_{J_k}$ (referenced in formula 10), generates the processed information from each target area. The target information is then computed as follows:

$$R_{jk} = \sum_{i=1}^{N_d} \sum_{j=1}^{d_n} F_{J_k}(d_{i,j}), \tag{11}$$

where $R_{jk}$ is the target result of information processing, $N_d$ and $d_n$ are the number of target areas in the respective dimensions, and $F_{J_k}(d_{i,j})$ is the function that processes information from area $(i,j)$.

Thus, the intelligent system $M_s$ is comprehensively defined across all levels of its architecture, according to formulas (8) and (9). This definition is essential for detailing both the system's components and the interactions among its various modules. As outlined in formula (8), the $M_s$ system operates as a distributed architecture, with modules integrated into software components as specified in formula (9). The architectural elements described in formula (8) coordinate and manage the components detailed in formula (9).

The architectural diagram of the $M_s$ system, shown in Figure 2, illustrates the modular level and the interactions between components.

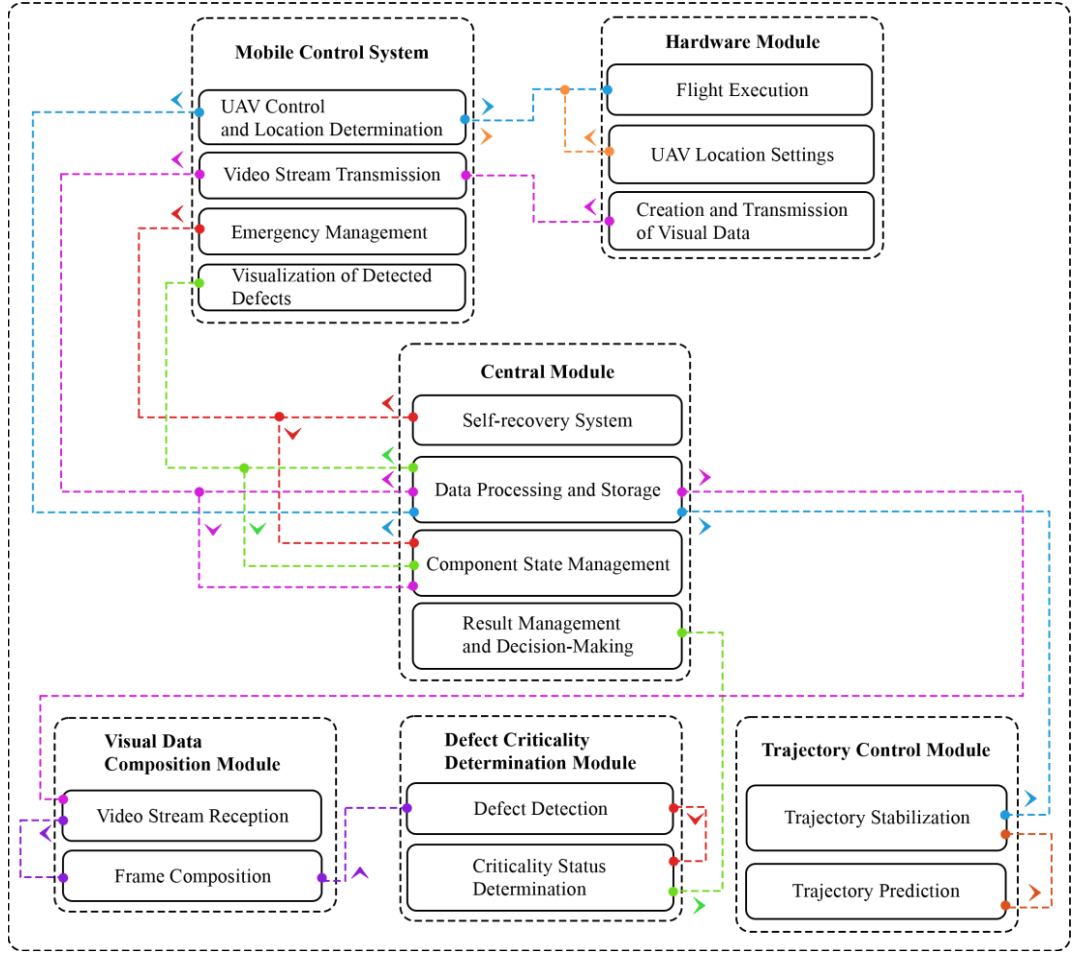

**Figure 2:** Modular-Level Architecture of the Intelligent System $M_s$.

Formulas (4)–(11) provide a precise definition of the $M_s$ system, detailing the variety of system components and their connections. The architecture presented at the modular level identifies specific modules within the software components, ensuring a system design that accommodates all potential complexities and enables component scalability through module reuse. Together, formulas (4)–(11) define the format of input and output data, providing a comprehensive description of the $M_s$ system. Further stages of $M_s$ system design require detailing each component according to these formulas.

The architecture shown in Figure 2 encompasses all components outlined in formulas (4)–(11), including the processes for generating and processing results, which constitute the system's output data. The main components of the system are as follows:

1. Mobile Control System, responsible for coordinating and monitoring UAVs. It performs the following key functions:

- Ensures accurate real-time determination of UAV coordinates and navigation in space.
- Transmits visual data collected by the UAV to the central module for further processing.
- Monitors and responds to emergency situations, ensuring the safety of UAV operations.
- Displays detected defects on an interactive interface.

2. Hardware Module, which integrates software components and provides:

- Implementation of flight algorithms and support for UAV stability in the air.
- Accurate determination and correction of UAV coordinates.
- Collection of visual information from cameras and other sensors and the transmission of this data to the mobile control system.

3. Central Module, a key element of the system, ensures:

- Continuous system operation even in the event of partial failures.
- Analysis and storage of information received from other modules.
- Monitoring the operational states of all system components, ensuring their optimal performance.
- Analysis of processed data and generation of recommendations or automatic decisions based on that data.

4. Visual Data Composition Module, which performs the following functions:

- Receives video data from the mobile control system.
- Processes and composes frames from various hardware sensors to provide high-quality visual analysis.

5. Defect Criticality Assessment Module, responsible for:

- Analyzing the data received to identify defects.
- Assessing the severity of detected defects and their impact on the object.

6. Trajectory Management Module, which performs the following tasks:

- Ensures the stability of the UAV's trajectory.
- Predicts future positions to optimize the UAV's route.

Therefore, the developed architecture of the intelligent $M_s$ system establishes the requirements for the key elements and components necessary for its operation, as well as defines the interconnections between them.

## 4. Results and discussion

For the experimental study of the effectiveness of the developed automated system $M_s$, the wind power plant "Staryi Sambir-1," [28] located in the Carpathian region of Ukraine, was selected. This station is equipped with Vestas V112 wind turbines, whose characteristics meet modern requirements for renewable energy equipment. Specifically, each wind energy unit (WEU) has a nominal power of 3 MW, a rotor diameter of 112 meters, and a tower height of 84 meters, providing significant area coverage of 9,852 m². Blades measuring 56 meters in length can operate efficiently in a wind speed range from 3 to 25 m/s, with a nominal speed of 12 m/s.

A comparison was conducted between the traditional manual approach and the automated intelligent system $M_s$ for data collection and processing to detect and evaluate the criticality of defects. Table 1 provides a comparative analysis of the key efficiency indicators for two methods: the traditional operator-based approach and the automated intelligent system $M_s$.

**Table 1**
Comparison of efficiency indicators between the traditional and automated $M_s$ data collection systems

| Method | Number of Simultaneous Inspections (N) | Inspection Time per WEU (hours) | Defect Detection (%) | Coverage Completeness (%) | Critical Determination Time (hours) | Data Update Frequency (months) |
|---|---|---|---|---|---|---|
| Traditional (operator) | 1 | 5.5–6.0 | 75–85 | 70–80 | 1.0–1.5 | 3 |
| System $M_s$ | 1 | 1.5–2.1 | 90–94 | 90–95 | 0.1–0.2 | 1 |
|  | 2 | 1.5–2.0 | 90–94 | 90–95 | 0.1–0.2 | 1 |

The conducted experiments demonstrate that the intelligent system notably improves defect detection accuracy, achieving 90–94%, and coverage completeness, reaching 90–95%, compared to the traditional method's 75–85% accuracy and 70–80% coverage.

Additionally, it can be seen in Table 1 that the automated system $M_s$ reduces the inspection time of a single WEU from 5.5–6.0 hours to 1.5–2.0 hours, enhancing the processing intensity of large data volumes and allowing regular updates on the WEU's condition (monthly, unlike the traditional method, which updates data every three months). Figure 3 illustrates the overall effectiveness of the intelligent system $M_s$ compared to the traditional data collection method.

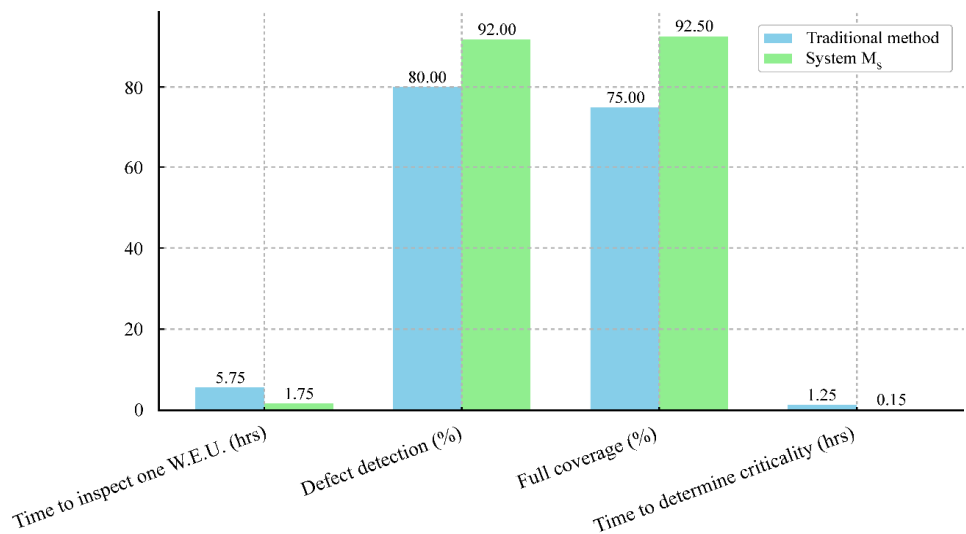

**Figure 3:** Illustration of the effectiveness of the automated intelligent system $M_s$ compared to the traditional data collection method.

Figure 3 visually underscores the advantages of the intelligent system by illustrating the increased precision and reduced operator dependence. The use of UAVs minimizes human error and safety risks, particularly for components located in challenging or hazardous environments, such as high-altitude turbine blades. By enabling regular inspections with higher accuracy and coverage, the system contributes to extending the operational lifespan of wind turbines and optimizing their performance.

Table 2 shows how the intelligent system $M_s$ provides more accurate and faster determination of defect criticality.

**Table 2**

Comparison of efficiency indicators between the traditional and automated $M_s$ data collection systems

| Defect Type | Method | Defect Size (cm) | WEU Component | Critical Assessment (1–10) | Critical Assessment Time (hours) |
|---|---|---|---|---|---|
| Crack | Traditional (manual) | 5.0–7.0 | Blade | 7–9 | 1.0–1.5 |
|  | System $M_s$. | 4.5–7.5 | Blade | 6–7 | 0.1–0.2 |
| Corrosion | Traditional (manual) | 10.0–15.0 | Tower | 7–9 | 1.0–1.5 |
|  | System $M_s$. | 9.0–14.0 | Tower | 8–9 | 0.1–0.2 |
| Overheating | Traditional (manual) | – | Generator | 7–9 | 1.0–1.5 |
|  | System $M_s$. | – | Generator | 6–7 | 0.1–0.2 |

For example, for blade cracks, the system reduces the criticality assessment time from 1.0–1.5 hours to 0.1–0.2 hours. This approach significantly enhances precision, allowing for prompt criticality assessments of defects such as tower corrosion and generator overheating, thereby reducing delays in decision-making regarding necessary repairs.

The automated system also significantly reduced inspection time, as shown previously in Table 1. The average inspection duration per WEU decreased from 5.5–6.0 hours using traditional methods to 1.5–2.0 hours with the intelligent system. This reduction enhances operational efficiency and facilitates more frequent inspections, allowing for the timely identification and mitigation of potential issues.

In addition to quicker inspections, Table 2 highlights the system's ability to reduce the time required for criticality assessments. For example, blade cracks were assessed within 0.1–0.2 hours using the automated system, compared to 1.0–1.5 hours with manual methods. Similar improvements were observed for other defect types, such as tower corrosion and generator overheating.

Figure 4 illustrates the narrowing of the criticality assessment range when using the intelligent system. Traditional methods, which rely heavily on subjective judgment, produced broader criticality ranges (e.g., 7–9 for blade cracks). In contrast, the intelligent system leveraged data-driven algorithms to achieve more precise assessments (e.g., 6–7 for the same defect type).

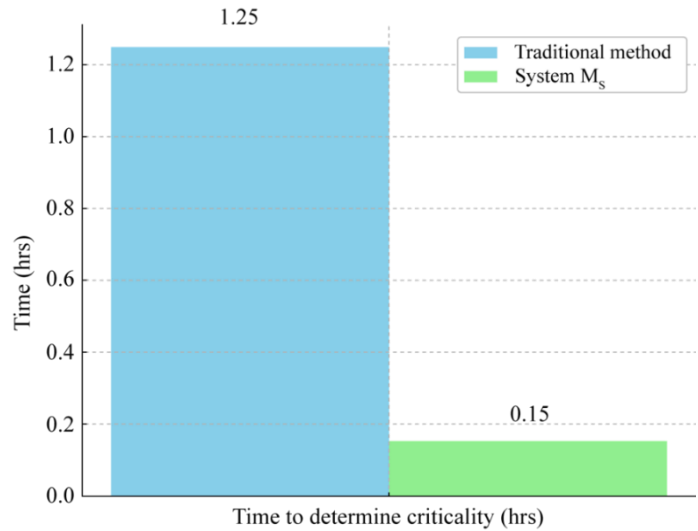

**Figure 4:** Results of defect criticality assessments using traditional and intelligent methods.

The experimental results also demonstrate the system's potential to improve safety and reduce operational costs. By minimizing the need for manual inspections in hazardous environments, the system mitigates risks to personnel while achieving better inspection outcomes. Moreover, the automation of data collection and processing reduces dependency on highly skilled operators, making the solution more accessible and cost-effective.

Despite its advantages, the proposed system has certain limitations that warrant further investigation. For instance, the reliance on advanced algorithms and high-quality sensors may pose challenges in terms of initial deployment costs and computational requirements. Additionally, adverse weather conditions, such as strong winds or poor visibility, could affect UAV performance and data accuracy. Addressing these challenges through algorithmic optimizations and robust hardware design will be critical for ensuring the system's reliability under diverse operating conditions.

## 5. Conclusions

The proposed intelligent system architecture greatly enhances the dynamic collection and analysis of visual data for detecting defects in wind turbines. Experimental results from the "Staryi Sambir-1" wind power plant show that the system boosts defect detection accuracy to 90–94%, compared to 75–85% achieved with traditional manual inspections. The inspection time per WEU is reduced from approximately 6 hours to as little as 1.5 hours, with coverage completeness increasing to 90–95%. Additionally, the system enables faster criticality assessments, cutting evaluation time from up to 1.5 hours to just 0.1–0.2 hours.

Despite these advancements, the proposed approach has limitations, such as the initial costs of deploying UAVs and sensors, the need for sophisticated algorithms to manage multi-sensor data fusion, and potential challenges in coordinating UAVs under adverse weather conditions. Moreover, reliance on high-quality data processing infrastructure may pose accessibility challenges in remote locations.

Future research will focus on mitigating these limitations by optimizing the system's cost-effectiveness, enhancing algorithmic performance under diverse environmental conditions, and refining UAV coordination mechanisms.